\documentclass[letterpaper]{article} 
\usepackage{aaai24}  
\usepackage{times}  
\usepackage{helvet}  
\usepackage{courier}  
\usepackage[hyphens]{url}  
\usepackage{graphicx} 
\usepackage{natbib}  
\usepackage{caption} 
\usepackage{newfloat}
\usepackage{listings}
\usepackage{bibentry}
\usepackage{algorithm}
\usepackage{algorithmic}
\usepackage{booktabs}       
\usepackage{amsfonts}       
\usepackage{nicefrac}       
\usepackage{microtype}      
\usepackage{xcolor}         
\usepackage{enumitem}
\usepackage{makecell}
\usepackage{amssymb}
\usepackage{amsthm}
\usepackage[switch]{lineno}
\usepackage{amsmath} 
\usepackage{subfigure}
\usepackage{multirow}
\usepackage{colortbl}
\usepackage{algorithm}
\usepackage{algorithmic}
\usepackage{booktabs}
\usepackage{url}
\usepackage{amsmath} 
\usepackage{subfigure}
\usepackage{multirow}

\usepackage{colortbl}

\definecolor{gray}{RGB}{222,222,222}
\DeclareCaptionStyle{ruled}{labelfont=normalfont,labelsep=colon,strut=off} 
\floatstyle{ruled}
\newfloat{listing}{tb}{lst}{}
\floatname{listing}{Listing}
\title{ Exploring One-Shot Semi-supervised Federated Learning\\with Pre-trained Diffusion Models}
\author {
    Mingzhao Yang \equalcontrib,
    Shangchao Su \equalcontrib,
    Bin Li \thanks{Corresponding author},
    Xiangyang Xue \footnotemark[2]
}
\affiliations {
Shanghai Key Laboratory of Intelligent Information Processing\\
School of Computer Science, Fudan University\\
\{mzyang20,scsu20,libin,xyxue\}@fudan.edu.cn 
}

\begin{document}

\maketitle

\begin{abstract}

Recently, semi-supervised federated learning (semi-FL) has been proposed to handle the commonly seen real-world scenarios with labeled data on the server and unlabeled data on the clients. However, existing methods face several challenges such as communication costs, data heterogeneity, and training pressure on client devices. To address these challenges, we introduce the powerful diffusion models (DM) into semi-FL and propose \textbf{FedDISC}, a \textbf{Fed}erated \textbf{D}iffusion-\textbf{I}nspired \textbf{S}emi-supervised \textbf{C}o-training method. Specifically, we first extract prototypes of the labeled server data and use these prototypes to predict pseudo-labels of the client data. For each category, we compute the cluster centroids and domain-specific representations to signify the semantic and stylistic information of their distributions. After adding noise, these representations are sent back to the server, which uses the pre-trained DM to generate synthetic datasets complying with the client distributions and train a global model on it. With the assistance of vast knowledge within DM, the synthetic datasets have comparable quality and diversity to the client images, subsequently enabling the training of global models that achieve performance equivalent to or even surpassing the ceiling of supervised centralized training. FedDISC works within one communication round, does not require any local training, and involves very minimal information uploading, greatly enhancing its practicality. Extensive experiments on three large-scale datasets demonstrate that FedDISC effectively addresses the semi-FL problem on non-IID clients and outperforms the compared SOTA methods. Sufficient visualization experiments also illustrate that the synthetic dataset generated by FedDISC exhibits comparable diversity and quality to the original client dataset, with a neglectable possibility of leaking privacy-sensitive information of the clients.

\end{abstract}

	\section{Introduction}
        \label{1}
 
    Federated Learning (FL)~\cite{mcmahan2017communication} is a new paradigm of machine learning that allows multiple clients to perform collaborative training without sharing private data. Realistic FL scenarios, such as mobile album classification and autonomous driving~\cite{nguyen2022deep,fantauzzo2022feddrive}, often involve individual users who are unwilling or unable to provide reliable annotations. This often results in the client data being unlabeled in practice. Semi-supervised FL (Semi-FL)~\cite{zhang2021improving,diao2021semifl,DBLP:conf/iclr/JeongYYH21} has been proposed to address this issue. In this setting, there are multiple clients with unlabeled data and a server with labeled data. The goal of semi-FL is to obtain a global model that adapts to all client distributions. 
    

	Due to its allowance for unlabeled client data, semi-FL should be the most practically valuable topic within FL. However, existing semi-FL methods are unable to be practically deployed in real-world scenarios due to the following reasons ~\cite{li2020federated,kairouz2021advances,mammen2021federated}: Firstly, the primary challenge lies in communication. Currently, all semi-FL methods heavily rely on multi-round communication,  which significantly increases the burden on the clients. Secondly, the challenge of data heterogeneity persists in semi-FL. When there are distribution differences between the server and the clients, the performance of the global model significantly decreases. The third point pertains to the diverse devices of clients in real-world scenarios. While many FL methods do not restrict the computing power of the clients, one of the greatest challenges in the practical implementation of FL methods is that most clients cannot support model training on their devices, such as a significant portion of mobile terminals in scenarios like mobile album classification and autonomous driving. Hence, to address the aforementioned challenges, it is essential to establish a \textbf{one-shot semi-FL method without any client training}.
  
    Recently, the development of diffusion models (DM)~\cite{radford2021learning,rombach2022high} offers fresh opportunities.  These pre-trained DMs exhibit remarkable performance. With proper guidance, these DMs can generate data with sufficient variety in both categories and distributions. If there is a method to generate guidance about the personalized distributions of the clients, it becomes possible to limitlessly generate high-quality, large-scale realistic images that comply with various client distributions in semi-FL. With the synthetic datasets, one can achieve one-shot semi-FL without the need for any client training, even in scenarios with highly non-IID clients. This approach simultaneously addresses the aforementioned three challenges and significantly enhances the practicality of semi-FL.
    
    Furthermore, an additional pivotal advantage of applying pre-trained DMs in semi-FL is the potential to surmount the "performance ceiling" of traditional FL, which involves uploading all client images to the server for centralized training of the global model. This ceiling bypasses any performance losses caused by distributed training and privacy preservation, enabling the training of the optimal global model. In semi-FL, this ceiling becomes even more unattainable, as the data uploaded to the server can be labeled, introducing additional supervised information. But even this ceiling performance is constrained by the knowledge within the client samples.
    However, with the vast knowledge within the DMs, it becomes possible to generate samples with both higher diversity and quality than the original client data, with the great possibility of surpassing the ceiling performance of centralized training.

Motivated by these opportunities, in this paper, we introduce \textbf{FedDISC}, a \textbf{Fed}erated \textbf{D}iffusion-\textbf{I}nspired \textbf{S}emi-supervised \textbf{C}o-training method, to leverage powerful foundation models in one-shot semi-FL. In brief, FedDISC involves four key steps: Firstly, following the common approach in semi-FL, we obtain prototypes for each category at the server and then send these prototypes to the clients. Secondly, the clients extract the features of the unlabeled client images and employ the received prototypes to assign pseudo-labels to these images. Thirdly, we obtain some cluster centroids and a domain-specific representation by clustering and averaging the client features for each category. These selected features possess the capability to capture the semantic information and the style characteristics of the personalized client distribution, which is then transmitted to the server. Finally, guided by the received representations, the server utilizes the pre-trained DM to limitlessly generate various samples complying with the specified distributions, resulting in a high-quality synthetic dataset. With the powerful DM, the generated samples closely resemble both the distribution and quality of the client dataset, enabling training a global model that achieves performance comparable to the ceiling performance or even surpassing it in some cases.

	Our experiments on DomainNet~\cite{peng2019moment}, OpenImage~\cite{kuznetsova2020open}, and NICO++~\cite{zhang2022NICO++} demonstrate that FedDISC can obtain a high-performance global model that adapts to various client distributions within one round communication. In some cases, it even outperforms the ceiling of centralized training. A large number of visualization experiments also demonstrate that we can generate synthetic datasets that exhibit quality and diversity comparable to the original client datasets, without the leaking of privacy-sensitive information. In certain cases, these synthetic datasets can even possess a more comprehensive knowledge than the original client datasets.
 
Our contributions are summarized as follows: 
\begin{itemize}
\item We demonstrate the excellent performance of DMs when applied to FL, enabling us to obtain high-quality large-scale synthetic datasets complying with the various client distributions without any training on the clients, which has not been explored before.
\item
We propose the FedDISC method. With the help of cluster centroids and domain-specific representations, our method further improves both the quality and variety of the generated samples, resulting in a global model that has the potential to outperform the performance ceiling of centralized training with only one communication round.
\item
We conduct extensive experiments on multiple real-world large-scale image datasets to validate the effectiveness of FedDISC. The results demonstrate that FedDISC outperforms compared with all the baseline methods. In some cases, it even surpasses the performance ceiling of traditional FL. The sufficient visualization experiments also illustrate that our method can generate synthetic datasets with competitive quality and diversity compared to the original client images, without leaking the privacy-sensitive information of the clients.

\end{itemize}

 \begin{figure*}[t]
 \centering
 \includegraphics[width=\linewidth]{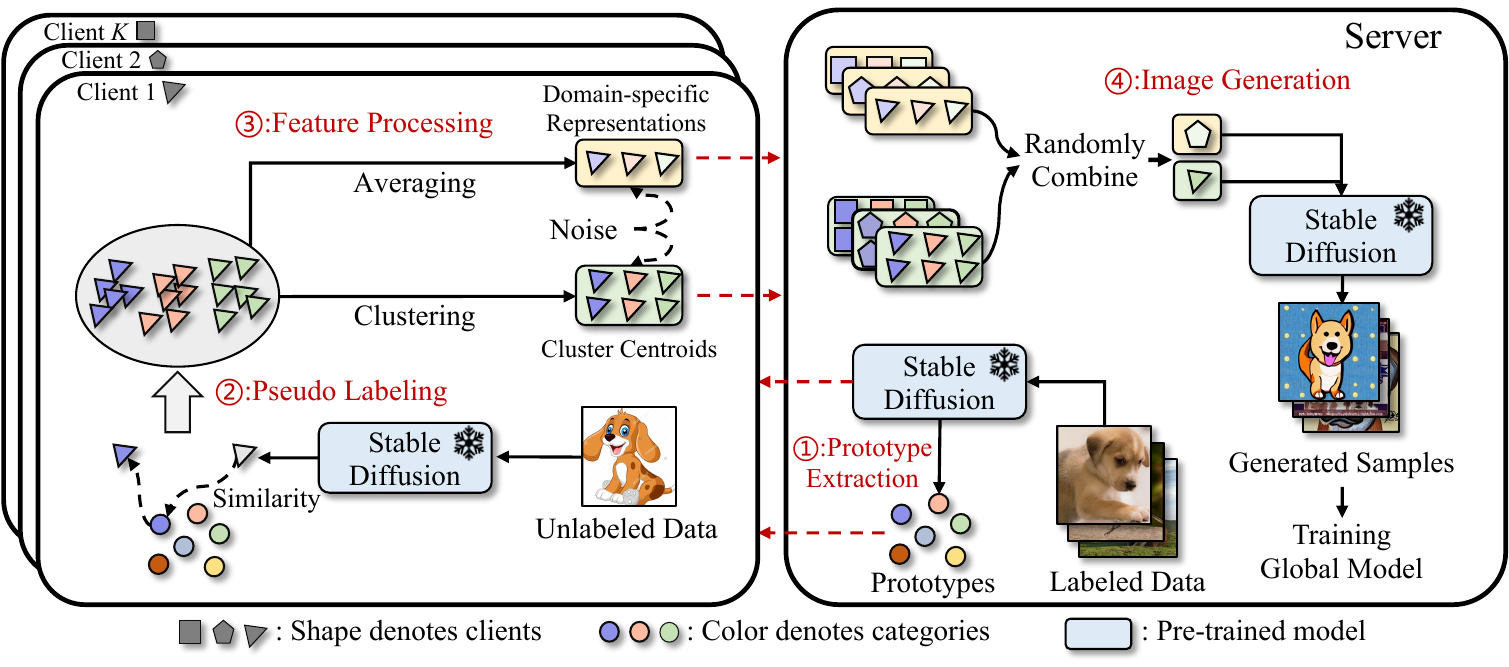}
 \centering
 \caption{
The framework of FedDISC. The overall method consists of four steps: Prototype Extraction, Pseudo Labeling, Feature Processing, and Image Generation.
 }
 \label{framework}
 \end{figure*}
 
\section{Related Works}
\subsection{Federated Learning}
\textbf{Supervised Federated Learning.} FedAvg~\cite{mcmahan2017communication} proposes the FL problem setting. However, some studies~\cite{li2019convergence,conf/mlsys/LiSZSTS20} notice the problem in non-IID scenarios. To address this challenge, numerous works have attempted to establish stronger global models ~\cite{karimireddy2020scaffold,conf/mlsys/LiSZSTS20,wang2020tackling,reddi2020adaptive}, or personalized FL that allows clients to obtain personalized parameters~\cite{fallah2020personalized,huang2021personalized,caruana1997multitask,DBLP:conf/nips/DinhTN20,DBLP:conf/iclr/ZhangSFYA21}. Some works~\cite{li2007support} involve aggregating distributed information from clients by uploading it to the server. In addition, to further reduce communication costs, some works~\cite{zhang2022dense,heinbaughdata,SU2023} propose one-shot FL, which performs one round of communication. 

 \textbf{Semi-supervised Federated Learning.}   In realistic FL scenarios, clients own a large amount of unlabeled data. In response to this issue, FedMatch~\cite{DBLP:conf/iclr/JeongYYH21} proposes semi-FL, which is mentioned in the introduction. ~\cite{zhang2021improving} points out the importance of the gradient diversity problem and proposes several strategies. ~\cite{diao2021semifl} supposes the additional auxiliary dataset to handle semi-FL. In the paper, an additional challenge about communication and client training is identified and a solution is proposed, enhancing the practicality of semi-FL.
 
\subsection{Foundation Models}

    Recently, foundation models~\cite{radford2021learning,kirillov2023segment,rombach2022high,yu2023chinese} have achieved unprecedented success in computer vision. CLIP~\cite{radford2021learning} has bridged the gap between text and vision. DMs~\cite{sohl2015deep,ho2020denoising} have provided a new generative paradigm, with Stable Diffusion~\cite{rombach2022high} achieving remarkable performance. A major advantage of DMs is the ability to use various conditions to guide generation, such as a trained classifier~\cite{dhariwal2021diffusion}, text~\cite{nichol2021glide,saharia2022photorealistic,kim2022diffusionclip,jin2023training} and images~\cite{saharia2022palette,zhang2023adding,su2022dual,wang2022pretraining,preechakul2022diffusion}. There are also some works~\cite{liu2022compositional,huang2023composer,du2023reduce} studying compositional generation of DMs.  From their performance in image generation, a large-scale pre-trained DM can generate realistic images within an acceptable cost of time and computation on the server. This is the main reason why we apply pre-trained DMs in FL.

\subsection{FL with Foundation Models}
Leveraging the powerful performance of fundamental models, some works~\cite{su2022cross,guo2022promptfl,yang2023one} in FL have explored the application of foundation models in federated image classification. However, to the best of our knowledge, no work has yet utilized pre-trained DMs, such as Stable Diffusion, in semi-FL. In this paper, we make a novel attempt and reveal the potential of DMs in semi-FL. Therefore, the method proposed in this paper, which does not require client training, has great potential for practical applications.

    \section{Method}
    \label{3}

    In this section, we introduce the proposed FedDISC method in detail. Firstly, we provide some notations and background knowledge on DMs. Then we describe the proposed method in detail through four steps taken by the clients and the server.
    

    \subsection{Preliminaries}
    \label{3.1}

    \textbf{Diffusion Models.} The DMs study the transformation from the Gaussian distribution to the realistic distribution by iterative denoising. In this paper, since only the pre-trained DMs are used and no training is conducted, the sampling process of the DMs is mainly introduced here. During sampling, the DM $\epsilon_{\theta}$ samples ${\mathbf{s}}_T$ from the Gaussian distribution, where $T$ is the predetermined maximum timestep. The DM takes ${\mathbf{s}}_T$ as the initial noise of the denoising process and uses the input text prompt $p$ and the input image $q$ as conditions. After $T$ timesteps of denoising, ${\mathbf{s}}_T$ is restored to a real image ${\mathbf{s}}_0$ with specified semantics. For any given time step $t\in\left \{ 0,\dots ,T \right \} $, the sampling process is as follows:
    \begin{linenomath}\begin{align}
    	\nonumber \mathbf{s}_{t-1} = &\sqrt{\alpha_{t-1}}\Big(\frac{\mathbf{s}_{t}-\sqrt{1-\alpha_{t}}\epsilon_{\theta}(\mathbf{s}_{t},t|p,q)}{\sqrt{\alpha_{t}}}\Big)\\
    	&+\sqrt{1-\alpha_{t-1}-\sigma_{t}^{2}}\cdot\epsilon_{\theta}(\mathbf{s}_{t},t|p,q)+\sigma_{t}\varepsilon _{t}
    \end{align}\end{linenomath}
    where $\alpha_{t}$, $\alpha_{t-1}$ and $\sigma_{t}$ are pre-defined parameters, $\varepsilon_{t}$ is the Gaussian noise randomly sampled at each timestep. It should be noted that currently, many methods can freely control the number of iterations of the denoising process to accelerate sampling, but the overall process is quite similar, so these methods won't be elaborated here.

    \textbf{Notations and Objectives.} We consider a semi-FL setting, where we have $K$ clients with unlabeled datasets $\mathcal{D}_k=\{\mathbf{x}_i^k\}_{i=1}^{N_k}, k=1,\ldots,K$, where ${N_k}$ is the number of images on the $k$-th client, and a server with a labeled dataset $\mathcal{D}_s=\{\mathbf{x}_i^s, y_i\}_{i=1}^{N_s}$, $y_{i} \in \left \{ 1,\ldots,M \right \}$, where ${N_s}$ is the number of images on the server and $M$ is the number of categories. The text prompts of these categories are $\mathcal{C}_j,j\in \left \{ 1,\ldots,M \right \}$. The objective of the whole FL framework is:
    \begin{linenomath}\begin{align}
        \min _{\mathbf{w} \in \mathbb{R}^{d}} \frac{1}{K} \sum_{k=1}^{K}  \mathbb{E}_{\mathbf{x} \sim \mathcal{D}_{k}}\left[\ell_{k}(\mathbf{w} ; \mathbf{x})\right] 
    \end{align}\end{linenomath}
where $\ell_{k}$ is the local objective function for the $k$-th client, $\mathbf{w}$ is the parameters of the global model.

To reduce communication and computation costs and make it suitable for real-world scenarios, such as the devices in autonomous driving, we impose two constraints in this setting: 1) Clients cannot conduct model training and can only conduct model inference. 2) The federated training process can only involve one round of communication.

    \subsection{FedDISC}
    \label{3.2}

	Our method has four detailed steps: prototype extraction, pseudo labeling, feature processing, and image generation.

    \textbf{Prototype Extraction.} 
    Firstly, we utilize a pre-trained CLIP image encoder $E_{\theta}$ to extract the features of all labeled data on the server. We assume that the server contains all possible categories that may appear on the clients, but each category on the server has a relatively single style or belongs to a limited fine-grained subclass.

    After obtaining the features of the labeled images on the server, we extract prototypes $\mathbf{p}_{j},j\in\left \{ 1,\dots ,M \right \} $ of all categories by calculating the average of all features with the same category:
    \begin{linenomath}\begin{align}
    \label{proto}
		\mathbf{p}_{j}=\frac{\sum_{(\mathbf{x}_i^s,y_i) \in \mathcal{D}_{s}} E_{\theta}(\mathbf{x}_i^s)*\mathbb{I}\left(y_{i}=j\right)}{\sum_{(\mathbf{x}_i^s,y_i) \in \mathcal{D}_{s}} \mathbb{I}\left(y_{i}=j\right)}
    \end{align}\end{linenomath}
    where $\mathbb{I}$ is the indicator function. Finally, we send the extracted category prototypes $\mathbf{p}_{j},j\in\left \{ 1,\dots ,M \right \} $ and the pre-trained CLIP image encoder $E_{\theta}$ to all the clients.

	\textbf{Pseudo Labeling.} 
 For client $k$, after receiving the encoder  $E_{\theta}$ and the prototypes $\mathbf{p}_{j}$ from the server, the client uses $E_{\theta}$ to extract features of all unlabeled images in $\mathcal{D}_k$ and calculates the similarities between each feature $E_{\theta}(\mathbf{x}_i^{k})$ and all category prototypes $\mathbf{p}_{j},j\in\left \{ 1,\dots ,M \right \} $.
    \begin{linenomath}\begin{align}
    	sim(E_{\theta}(\mathbf{x}_i^{k}),\mathbf{p}_{j}) = \frac{E_{\theta}(\mathbf{x}_i^{k})^{\top}\mathbf{p}_j}{\left \| E_{\theta}(\mathbf{x}_i^{k}) \right \|\left \| \mathbf{p}_j \right \| }, \mathbf{x}_i^{k}\in \mathcal{D}_k
    \end{align}\end{linenomath}		
	
	Based on the similarities, each image $\mathbf{x}_i^{k}$ is assigned with a pseudo-label $\hat{y}_i^{k}$, where $\hat{y}_i^{k} = \mathop{\arg\max}_{j}sim(E_{\theta}(\mathbf{x}_i^{k}),\mathbf{p}_{j}) $. Due to the differences between $\mathcal{D}_s$ and $\mathcal{D}_k$, there is a possibility of making mistakes in pseudo labeling. In traditional semi-FL methods, pseudo-labels are used for self-training.  Therefore, various semi-FL methods are required to improve the quality of pseudo-labels. However, in our method, on one hand, in feature processing, clustering can avoid uploading representations of incorrect categories. On the other hand, the introduction of text prompts during the generation process can also prevent the generation of images that do not correspond to the specified categories, which further ensures the correct semantic information of generated images.

	\textbf{Feature Processing.} 
 After obtaining the unlabeled client features and their pseudo labels $\{ E_{\theta}(\mathbf{x}_i^{k}),\hat{y}_i^{k}\}_{i=1}^{N_k}$, taking category $j$ as an example, we cluster the client features belonging to category $j$ and select $L$ cluster centroids $\{\mathbf{z}_{j,l}^{k}\}_{l=1}^L$ to upload. The objective of clustering is as follows:
	\begin{linenomath}\begin{align}
	\label{cluster}
		\underset{\mathbf{z}_{j,l}^{k}}{\arg \min } \sum_{l=1}^{L} \sum_{\mathbf{x}_i^k\in\mathcal{D}_k}\left\|E_{\theta}(\mathbf{x}_i^{k})-\mathbf{z}_{j,l}^{k}\right\|^{2}*\mathbb{I}\left(\hat{y}_{i}^k=j\right)
	\end{align}\end{linenomath}	

	Compared with randomly selecting $L$ features for uploading, the personalized distribution information and semantic information contained in the cluster centroids are clearer. Since only a small number of features are selected and each feature will generate multiple images on the server, the quality of the uploaded features is crucial.

	Meanwhile, we obtain the domain-specific representations $\{\mathbf{g}_j^{k}\} _{j=1}^{M}$ for each category on client $k$ by averaging all the features belonging to category $j$. We weaken the individuality of each image and highlight the commonality of each category on the clients. During the conditional generation on the server, we can generate images complying with different client distributions by combining the cluster centroids with different domain-specific representations.

    After computing the cluster centroids and the domain-specific representations, for privacy protection, we add noise to all these features. The noise-adding process is as follows:
    \begin{linenomath}\begin{align}
    \nonumber
		\bar{\mathbf{z}}_{j,l}^{k} = \sqrt{\alpha_n}{\mathbf{z}}_{j,l}^{k} +\sqrt{1-\alpha_n}\varepsilon_1,
        \bar{\mathbf{g}}_{j}^{k} = \sqrt{\alpha_n}{\mathbf{g}_j^{k}} +\sqrt{1-\alpha_n}\varepsilon_2
	\end{align}\end{linenomath}		
	 where $\varepsilon_1, \varepsilon_2 \sim \mathcal{N}(0,\mathcal{I})$, $n$ is a hyperparameter controlling the intensity of the noise, and $n \in \{0, \ldots, T \}$. We follow the noise-adding process in Stable Diffusion~\cite{rombach2022high} and perform a noise-adding process with a specific timestep to these image features. After this step, cluster centroids $\{\bar{\mathbf{z}}_{j,l}^{k}\}_{l=1}^L,j=\{1, \ldots,M\}$ and domain-specific representations $\{\bar{\mathbf{g}}_j^{k}\} _{j=1}^{M}$ are uploaded to the server.
\begin{table*}[]
\center
\large
\resizebox{\linewidth}{!}{
\begin{tabular}{c|cccccc|cccccc}
\Xhline{1pt} \rowcolor{gray}
\multicolumn{1}{c|}{}           &            & \multicolumn{5}{c|}{OpenImage}                                                     & \multicolumn{6}{c}{DomainNet}                                                  \\ \cline{2-13} 
 \rowcolor{gray}
\multicolumn{1}{c|}{}            & client0 & client1 & client2 & client3 & \multicolumn{1}{c|}{client4} & average & clipart & infograph & painting & quickdraw & \multicolumn{1}{c|}{sketch} & average \\ \Xhline{1pt}
\multicolumn{1}{c|}{\textit{Ceiling}}         &\textit{ 54.05} & \textit{58.42} & \textit{62.59} & \textit{63.21 }   & \multicolumn{1}{c|}{\textit{64.79}}  &\textit{ 60.61} & \textit{81.54} & \textit{52.49} &\textit{ 73.54} & \textit{30.11 } & \multicolumn{1}{c|}{\textit{72.34}}   & \textit{62.01} \\ \Xhline{1pt}
\multicolumn{1}{c|}{Fine-tune}     & 36.67 & 46.81 & 45.43 & 47.17      & \multicolumn{1}{c|}{42.1} & 43.64 & 67.57 & 45.47 & 65.28 & 10.42   & \multicolumn{1}{c|}{62.14}   & 50.17 \\
\multicolumn{1}{c|}{Zero-shot}      & 56.03    & 40.61 & 40.28 & 44.06    & \multicolumn{1}{c|}{\textbf{61.45}}  & 48.47  & 65.86 & 40.5  & 62.25 & 13.36   & \multicolumn{1}{c|}{57.92}   & 47.98   \\
\multicolumn{1}{c|}{Prompt}       & 48.61 & 54.03 & 59.07 & 58.42     & \multicolumn{1}{c|}{53.49}  & 54.72 & 66.42 & 37.45 & 59.62 & 10.73   & \multicolumn{1}{c|}{63.92}   & 47.63   \\  \Xhline{1pt}
\multicolumn{1}{c|}{FedAvg}     & 41.11 & 44.06 & 46.57 & 47.45    & \multicolumn{1}{c|}{37.63} & 43.36 & 49.95 & 30.67 & 51.07 & 1.74   & \multicolumn{1}{c|}{38.46}   & 34.38   \\
\multicolumn{1}{c|}{SemiFL}       & 48.15 & 52.78 & 61.05 & 55.23    & \multicolumn{1}{c|}{46.16} & 52.67 & 69.55 & \textbf{47.16} & 64.54 & 7.02   & \multicolumn{1}{c|}{63.32}   & 50.32    \\
\multicolumn{1}{c|}{RSCFed}     & 28.97 & 38.04 & 40.82 & 33.98    & \multicolumn{1}{c|}{36.35} & 35.63 & 71.5  & 45.73 & 61.96 & 11.53  & \multicolumn{1}{c|}{65.03}   & 51.15   \\ \Xhline{1pt}
\multicolumn{1}{c|}{FedDISC}        & \textbf{56.11} & \textbf{62.49} & \textbf{62.53} & \textbf{59.16}    & \multicolumn{1}{c|}{56.77}  & \textbf{59.42}  & \textbf{72.54} & 43.47 & \textbf{67.42} & \textbf{17.71}   & \multicolumn{1}{c|}{\textbf{67.25}}   &  \textbf{53.68}   \\ \Xhline{1pt}

              \rowcolor{gray}
\multicolumn{1}{c|}{}           &            & \multicolumn{5}{c|}{NICO++\_C}                                                     & \multicolumn{6}{c}{NICO++\_U}                                                  \\ \cline{2-13} 
 \rowcolor{gray}
               \rowcolor{gray}
\multicolumn{1}{c|}{}                                & client0 & client1 & client2 & client3 & \multicolumn{1}{c|}{client4}  & average & client0 & client1 & client2 & client3 & \multicolumn{1}{c|}{client4} & average \\ \Xhline{1pt}
\multicolumn{1}{c|}{\textit{Ceiling}}         & \textit{89.19} &\textit{ 91.9}  & \textit{89.51} & \textit{90.47}     & \multicolumn{1}{c|}{\textit{85.1}}  & \textit{89.23} & \textit{96.35} & \textit{96.42} & \textit{96.88} & \textit{97.01}   & \multicolumn{1}{c|}{97.26}   & \textit{96.78}  \\ \Xhline{1pt}
\multicolumn{1}{c|}{Fine-tune}      & 86.5  & 89.39 & 83.61 & 87.21      & \multicolumn{1}{c|}{76.95} & 84.73 & 84.75 & 79.08 & 81.48 & 86.58   & \multicolumn{1}{c|}{83.52}   & 83.08 \\
\multicolumn{1}{c|}{Zero-shot}    & 78.66 & 85.26 & 80.01 & 80.7    & \multicolumn{1}{c|}{72.14} & 79.35 & 89.2  & 89.24 & 87.19 & 85.5   & \multicolumn{1}{c|}{88.6}   & 87.94  \\
\multicolumn{1}{c|}{Prompt}           & 86.94 & 87.41 & \textbf{89.73} & 82.69   & \multicolumn{1}{c|}{73.51}  & 84.05 & 90.61 & 87.14 & \textbf{89.96} & 87.48 & \multicolumn{1}{c|}{88.16}   & 88.67  \\  \Xhline{1pt}
\multicolumn{1}{c|}{FedAvg}   & 86.98 & 90.82 & 82.68 & 87.57    & \multicolumn{1}{c|}{74.48} & 84.51 & 83.26 & 73.3  & 77.93 & 80.8   & \multicolumn{1}{c|}{79.28}   & 78.91   \\
\multicolumn{1}{c|}{SemiFL}       & 87.55 & 89.27 & 81.93 & 87.16    & \multicolumn{1}{c|}{77.01} & 84.58 & 78.21 & 74.7  & 79.87 & 80.69   & \multicolumn{1}{c|}{77.02}   & 78.09    \\
\multicolumn{1}{c|}{RSCFed}       & 52.08 & 60.15 & 52.6  & 55.35    & \multicolumn{1}{c|}{43.89} & 52.81 & 71.88 & 64.14 & 70.82 & 69.71 & \multicolumn{1}{c|}{69.67}   & 69.24  \\ \Xhline{1pt}
\multicolumn{1}{c|}{FedDISC}        & \textbf{87.97} & \textbf{92.09} & 86.44 & \textbf{90.52}    & \multicolumn{1}{c|}{\textbf{84.17}} & \textbf{88.24} & \textbf{91.73} & \textbf{90.82} & 89.63 & \textbf{92.83}   & \multicolumn{1}{c|}{\textbf{90.15}}   &  \textbf{91.03}   \\ \Xhline{1pt}
\end{tabular}}
\caption{The performances of different methods on OpenImage, DomainNet, and NICO++, where the italicized texts represent the inaccessible supervised ceiling performance used solely as a reference, and bold texts represent the best performance excluding the supervised ceiling performance.}
\label{MainResult}
\end{table*}
\begin{figure}[t]
 \centering
 \includegraphics[width=\linewidth]{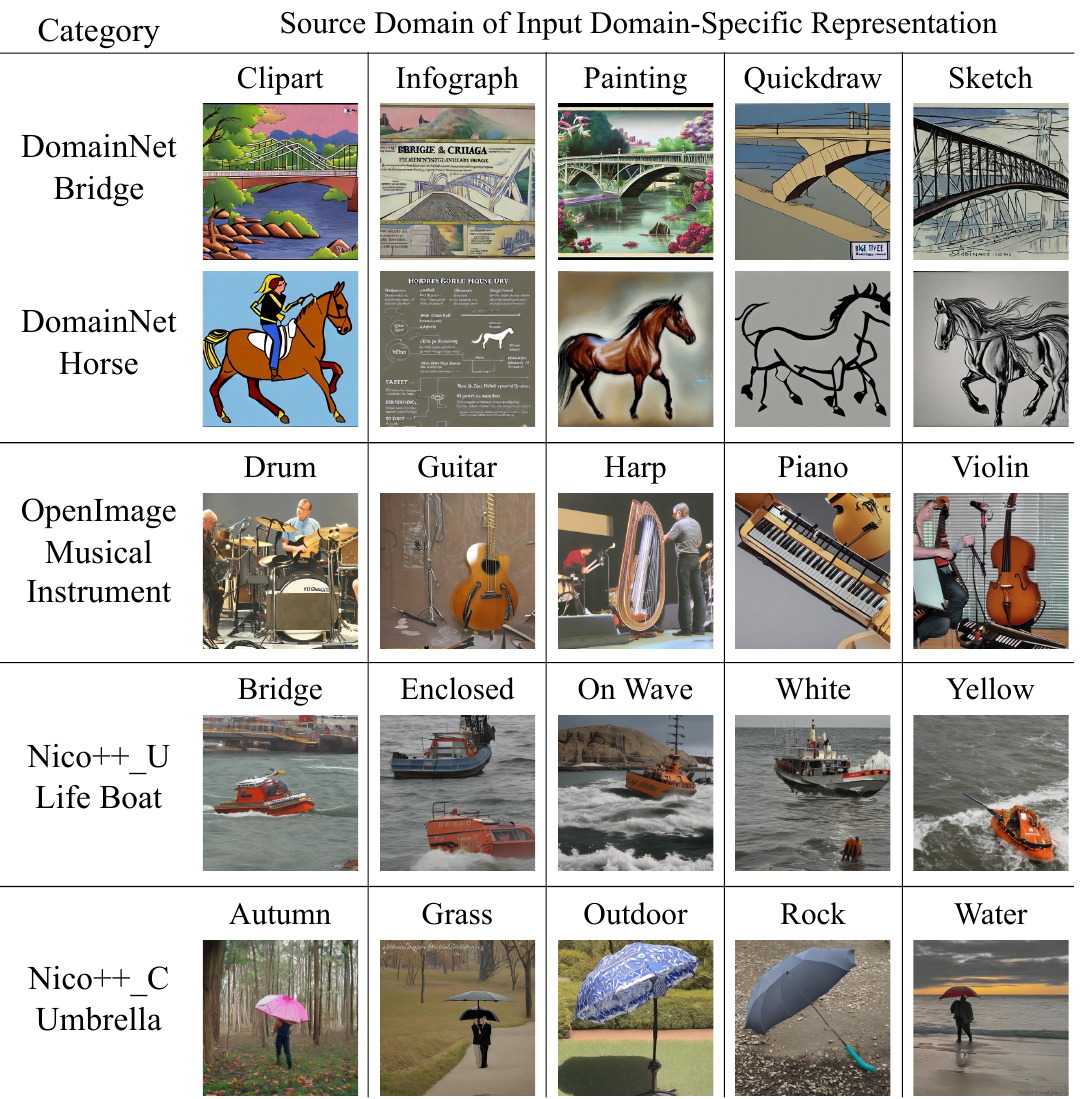}
 \centering
 \caption{
Generated images comply with different distributions on different datasets.
 }
 \label{MainVis}
 \end{figure} 

	\textbf{Image Generation.} 
 After receiving $\bar{\mathbf{z}}_{j,l}^{k}$ and $\bar{\mathbf{g}}_j^{k}$ uploaded from the clients, for each cluster centroid $\bar{\mathbf{z}}_{j,l}^{k}$, the server randomly combines $\bar{\mathbf{z}}_{j,l}^{k}$ with the domain-specific representations which have the same pseudo-label $j$, the selected domain-specific representations $\mathcal{G}_{j,l}^k = \{ \bar{\mathbf{g}}_j^{k_0}, \ldots,\bar{\mathbf{g}}_j^{k_R}\}$ is a random subset of domain-specific representations $\{\bar{\mathbf{g}}_j^{k}\} _{j=1}^{M}$.

	As for the generating process, following ~\cite{liu2022compositional}, since we aim to use the cluster centroids $\bar{\mathbf{z}}_{j,l}^{k}$, domain-specific representations $\bar{\mathbf{g}}_j^{k_i}$, and the text prompts $\mathcal{C}_j$ as conditions for generation, the conditional probability distribution of the sample $\mathbf{s}$ in diffusion process can be written in the following form:
	\begin{linenomath}
    \begin{align*}
	p(\mathbf{s}|\bar{\mathbf{z}}_{j,l}^{k},\bar{\mathbf{g}}_j^{k_i},\mathcal{C}_j)\propto p(\mathbf{s}|\mathcal{C}_j)p(\bar{\mathbf{z}}_{j,l}^{k}|\mathbf{s},\mathcal{C}_j)p(\bar{\mathbf{g}}_j^{k_i}|\mathbf{s},\mathcal{C}_j)
	\end{align*}
 \end{linenomath}

 Since $\mathbf{s}$ is initially sampled from a Gaussian distribution, independent of the used cluster centroids and domain-specific representations, the above formula can be rewritten as:
	\begin{linenomath}\begin{align*}
		p(\mathbf{s}|\bar{\mathbf{z}}_{j, l}^{k}, \bar{\mathbf{g}}_{j}^{k_{i}},\mathcal{C}_j) \propto p(\mathbf{s}|\mathcal{C}_j) \frac{p(\mathbf{s}|\bar{\mathbf{z}}_{j, l}^{k},\mathcal{C}_j)}{p(\mathbf{s}|\mathcal{C}_j)} \frac{p(\mathbf{s}|\bar{\mathbf{g}}_{j}^{k_{i}},\mathcal{C}_j)}{p(\mathbf{s}|\mathcal{C}_j)}
	\end{align*}\end{linenomath}

 Therefore, specifically, we use the feature of category prompt $\mathcal{C}_j$ with a cluster centroid $\bar{\mathbf{z}}_{c,l}^{k}$, a domain-specific representation $\bar{\mathbf{g}}_c^{k_i}$, and without any image feature to respectively obtain three predicted noises. We accumulate these three predicted noises in the following formula to obtain the final predicted noise:
	\begin{linenomath}\begin{align*}
		\hat{\epsilon}_{\theta}&(\mathbf{s}_t,t|\bar{\mathbf{z}}_{j,l}^{k},\bar{\mathbf{g}}_c^{k_i},\mathcal{C}_j)=\epsilon_{\theta}(\mathbf{s}_t,t|\mathcal{C}_j)+w_f(\epsilon_{\theta}(\mathbf{s}_t,t|\bar{\mathbf{z}}_{j,l}^{k},\mathcal{C}_j)\\
     &-\epsilon_{\theta}(\mathbf{s}_t,t|\mathcal{C}_j))+w_g(\epsilon_{\theta}(\mathbf{s}_t,t|\bar{\mathbf{g}}_j^{k_i},\mathcal{C}_j)-\epsilon_{\theta}(\mathbf{s}_t,t|\mathcal{C}_j))
	\end{align*}\end{linenomath}
	where $w_f$ and $w_g$ are the weights of the predicted noises. Overall, the generated images are obtained through the denoising process:
	\begin{linenomath}\begin{align}
	\label{generate}
	\nonumber
		& \mathbf{s}_{t-1}= \sqrt{\alpha_{t-1}}\Big(\frac{\mathbf{s}_{t}-\sqrt{1-\alpha_{t}}\hat{\epsilon}_{\theta}(\mathbf{s}_{t},t|\bar{\mathbf{z}}_{j,l}^{k},\bar{\mathbf{g}}_j^{k_i},\mathcal{C}_j)}{\sqrt{\alpha_{t}}}\Big)\\
		&+\sqrt{1-\alpha_{t-1}-\sigma_{t}^{2}}\cdot\hat{\epsilon}_{\theta}(\mathbf{s}_{t},t|\bar{\mathbf{z}}_{j,l}^{k},\bar{\mathbf{g}}_j^{k_i},\mathcal{C}_j)+\sigma_{t}\varepsilon_{t}
	\end{align}\end{linenomath}
	
 After obtaining the generated images, since both the cluster centroids and domain-specific representations used for image generation have their corresponding pseudo-labels, the generated images are pseudo-labeled. So we can directly fine-tune a classification model $h$ with the generated dataset for downstream classification tasks. The classification model $h = F_\theta \circ E_\theta$ is a composite of the pre-trained CLIP image encoder $E_\theta$ and a linear classifier $F_\theta$.

	\section{Experiments}
	\label{4}


    \subsection{Experimental Setup}
    \textbf{Datasets.}
    We adopt three datasets to evaluate the performance of FedDISC: DomainNet~\cite{peng2019moment}, OpenImage~\cite{kuznetsova2020open}, and NICO++~\cite{zhang2022NICO++}. NICO++ can be divided into the common contexts (NICO++\_C) and the unique contexts (NICO++\_U). We divide each dataset into six clients based on the inherent domain division of the dataset itself. Due to space constraints, we provide a comprehensive description of our dataset in the supplementary materials.
    
    \textbf{Compared Methods.}
     We mainly compare our method with 7 methods: 1) \textbf{Fine-tune}: Directly fine-tuning the model with the uploaded cluster centroids and corresponding pseudo-labels. 2) \textbf{Zero-shot}: Using the zero-shot classification capability of pre-trained CLIP to classify client images without any additional training. 3) \textbf{FedAvg}: Using clients' data and corresponding pseudo-labels for conducting FedAvg. In addition, we evaluate the SOTA semi-FL methods: 4) \textbf{SemiFL} ~\cite{diao2021semifl} and 5) \textbf{RSCFed}~\cite{liang2022rscfed}. Note that as there is currently no one-shot semi-FL method, the chosen semi-supervised methods require multiple rounds of communication for comparison. 6) \textbf{Prompts}: Image generation is directly performed without utilizing any image features as guidance and solely relying on the text prompts.  7) \textbf{Ceiling}. As mentioned in the introduction, the inaccessible performance ceiling involves directly uploading and labeling all client images to the server for training the aggregated model, also known as centralized training. And it needs to be mentioned that the \textbf{FedAvg}, \textbf{RSCFed}, and \textbf{SemiFL} all need multiple iterations.

   \begin{figure}[t]
 \centering
 \includegraphics[width=\linewidth]{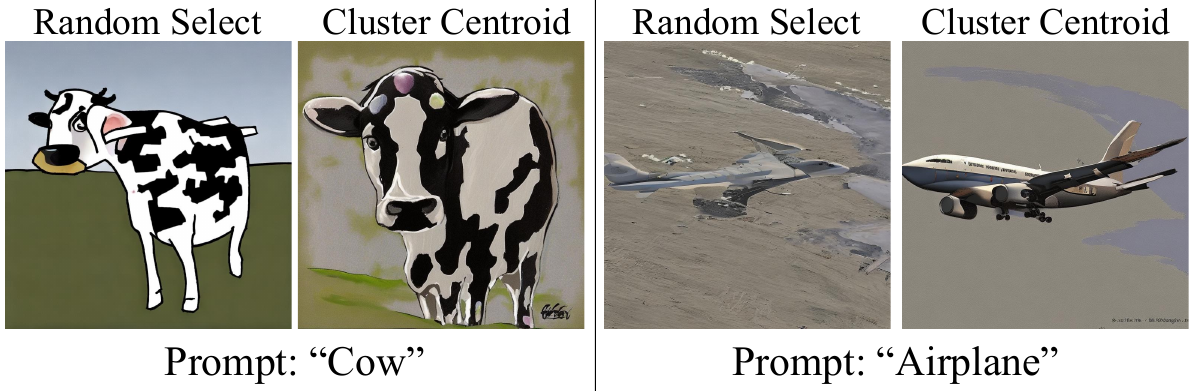}
 \centering
 \caption{
Comparison between generating using clustering centroids and the randomly selected client representations. With the provision of clustering centroids, the introduction of more representative semantic information leads to a significant improvement in the stability of the generated outputs.
 }
 \label{ClusterVis}
 \end{figure}	
 
 \begin{figure}[t]
 \centering
 \includegraphics[width=\linewidth]{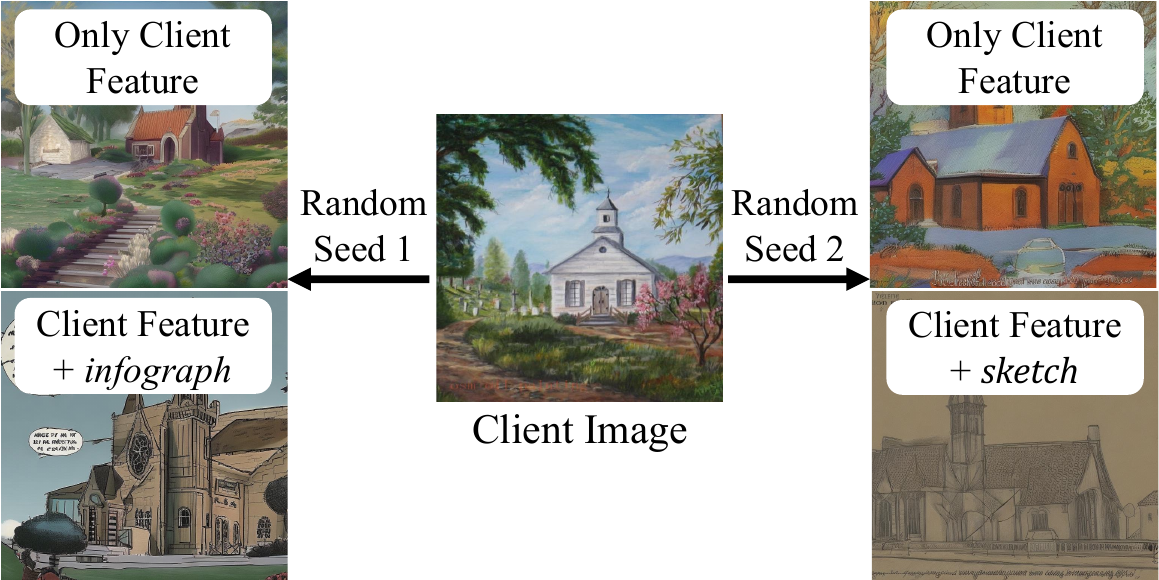}
 \centering
 \caption{
The inclusion of domain-specific representations and their impact on the generated results. We can effectively alter the style of the generated images by controlling the added domain-specific representations, thereby enhancing the diversity of generated samples.
 }
 \label{GlobalVis}
 \end{figure}
\begin{table}[t]
\center
\huge
\resizebox{1.0\linewidth}{!}{
\begin{tabular}{cccccc}
\toprule
 & client0 & client1 & client2 & client3 & client4  \\ \midrule
$L=3$ Fine-tune & 36.01 & 46.01 & 44.59 & 45.55 & 41.87  \\ 
$L=3$ FedDISC & \textbf{56.33} & \textbf{61.93} & \textbf{58.62} & \textbf{56.71} & \textbf{58.74}  \\ \midrule
$L=5$ Fine-tune & 36.67 & 46.81 & 45.43 & 47.17 & 42.10  \\ 
$L=5$ FedDISC & \textbf{56.11} & \textbf{62.49} & \textbf{62.53} & \textbf{59.16} & \textbf{56.77}  \\ \midrule
$L=10$ Fine-tune & 37.55 & 45.86 & 44.85 & 46.01 & 42.15  \\ 
$L=10$ FedDISC & \textbf{57.16} & \textbf{63.84} & \textbf{61.12} & \textbf{57.91 }& \textbf{59.13} \\ \bottomrule
\end{tabular}}
\caption{The influence of the number of cluster centroids.}
\label{centers}
\end{table}

\begin{table}[t]
\centering
\resizebox{\linewidth}{!}{
\begin{tabular}{ccccccc}
\toprule
 & \multicolumn{1}{c}{client0} & \multicolumn{1}{c}{client1} & \multicolumn{1}{c}{client2} & \multicolumn{1}{c}{client3} & \multicolumn{1}{c}{client4}  \\ \midrule
$R=3$ & 53.41 & 62.15 & 61.31 & 56.87 & 55.49  \\ 
$R=5$ & 54.58 & \textbf{63.47} & 61.26 & 58.19 & \textbf{57.57}  \\ 
$R=10$ & \textbf{56.11 }& 62.49 & \textbf{62.53} & \textbf{59.16} & 56.77  \\ \bottomrule
\end{tabular}}
\caption{The influence of the number of generated images.}
\label{gen_images}
\end{table}
 
\begin{table}[t]
\center
\resizebox{\linewidth}{!}{
\begin{tabular}{ccccccc}
\toprule
\begin{tabular}[c]{@{}c@{}}DR\end{tabular}  & \begin{tabular}[c]{@{}c@{}}CC\end{tabular} 
 & client0 & client1 & client2 & client3 & client4  \\ \midrule
 &  & 66.42 & 37.45 & 59.62 & 10.73 & 63.92  \\ 
$\checkmark$ &  & 67.79 & 40.02 & 63.59 & 13.77 & 60.57  \\ 
 & $\checkmark$ & 65.83 & 38.27 & 64.56 & 14.30 & 60.37 \\ 
$\checkmark$ & $\checkmark$ & \textbf{72.54} & \textbf{43.47} & \textbf{67.42} & \textbf{17.71} & \textbf{67.25}  \\ \bottomrule
\end{tabular}}
\caption{The influence of different conditions.}
\label{ablation}
\end{table}

\begin{table}[t]
\center
\huge
\resizebox{1.0\linewidth}{!}{
\begin{tabular}{ccc}
\toprule
     & Upload Params (M) & Client Compute (Gflops) \\ \midrule
    FedAvg     & 30*632.08 & 30*1004.19 \\
    SemiFL    & 500*632.08 & 500*1004.19 \\
    RSCFed    & 100*632.08 & 100*1004.19 \\
    Ceiling     & 925.88 & - \\
    FedDISC     & \textbf{4.23} & \textbf{334.73} \\ \bottomrule
\end{tabular}}
\caption{Comparison about communication and computation.}
\label{commu}
\end{table}

\subsection{Main Results}

Table~\ref{MainResult} shows the performance of our method and various compared methods on four datasets. We highlight several observations:
\begin{itemize}
\item In addition to the Ceiling, FedDISC achieves the best average performance with only 1 communication round. This demonstrates the potential of DMs in FL.
\item From the results on DomainNet, we can see that FedDISC has good performance on all clients except \textit{infograph}. A possible reason is that the Stable Diffusion still has limited support for text in the images currently. 
\item On OpenImage, NICO++\_C, and NICO++\_U, compared with other baselines, FedDISC exhibits a significant performance improvement. That's because Stable Diffusion is exposed to more realistic images during pre-training.
\item Despite the powerful ability of CLIP, directly using CLIP to perform classification still cannot achieve the best performance, therefore further fine-tuning is needed.
\item Compared with Prompts Only, without the guidance of client image features, generated images would be heavily biased towards the most common distributions, demonstrating the necessity of guidance.
\item Compared with Ceiling, FedDISC does not exhibit significant performance lag. It can even surpass in some domains, affirming the ideas posited in the introduction.

\end{itemize}

From Figure~\ref{MainVis} and the other visualization results in the supplementary materials, it can be seen that on DomainNet, OpenImage, NICO++\_C, and NICO++\_U our method can generate high-quality images that comply with various client distributions while being semantically correct, collectively underscoring the superior performance of FedDISC.

\subsection{Ablation Experiments}

	\textbf{The Number of Uploaded Cluster Centroids.} We perform experiments on OpenImage to discuss the influence of the number of the uploaded cluster centroids $L$. Since this number is related to the performance of fine-tuning, we also test the performance of fine-tuning under different $L$ for comparison. From Table~\ref{centers}, we can see that in most cases, uploading a small number of cluster centroids is already sufficient to represent the semantics of the subcategories. Increasing the number of cluster centroids enhances the availability of client information during the generation process, thereby further elevating the quality of generated images.

 \begin{figure}[t]
 \centering
 \includegraphics[width=\linewidth]{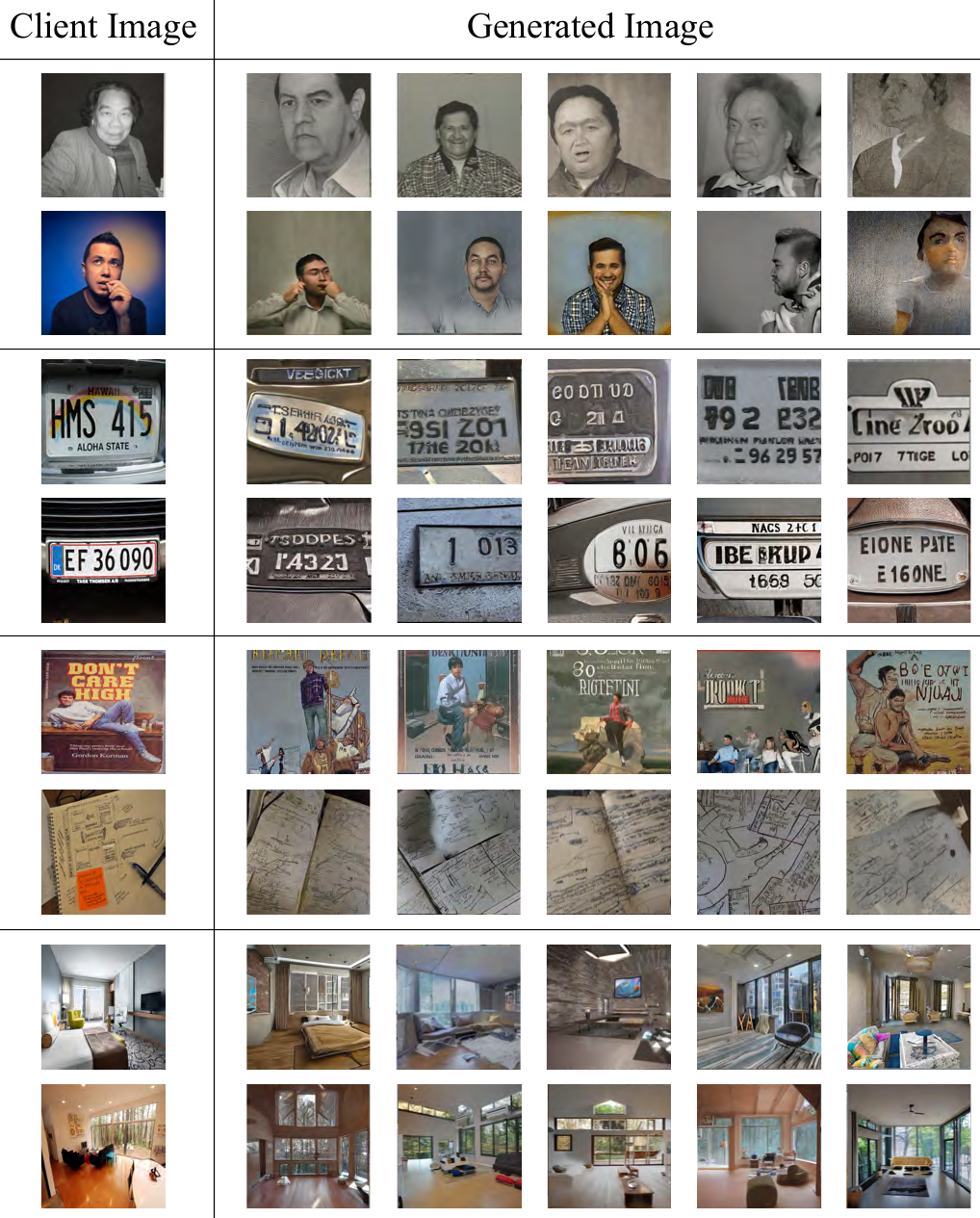}
 \centering
 \caption{
The comparison between the raw client images and their generated images. It can be observed that the generated images do not leak any sensitive privacy present in the original images, such as faces, text, etc. The generated images exhibit only a stylistic resemblance to the original images. Restoring original images starting from high-dimensional features without any training is nearly impossible.
 }
 \label{SupVis_privacy}
\end{figure}  
 
	\textbf{The Number of Generated Images.} We discuss the number of images generated by each cluster centroid on OpenImage. From Table~\ref{gen_images}, we can find that the overall performance of the method gradually improves as $R$ increases. This is reasonable as the increment of $R$ indicates the increment in the variety of generated data. However, generating a small number of images is sufficient since the randomly sampled initial noise can bring some varieties as well.

	\textbf{The Roles of Domain-specific Representations and Cluster Centroids.} We discuss the roles of Domain-specific Representations (DR) and Cluster Centroids (CC) on DomainNet. We compare FedDISC with cases where DR are not used during generation, and cases where CC are not uploaded, but an equal number of client features are randomly uploaded. As shown in Table~\ref{ablation}, the removal of either DR or CC has a significant influence on the performance. 

    The visualization results in Figure~\ref{ClusterVis} demonstrate that without the cluster centroids, the semantic information of the generated images becomes ambiguous and may lead to generating images that do not match the given text prompts. Figure~\ref{GlobalVis} shows that removing the domain-specific representations leads to the style of the generated images being monotonous, which reduces the diversity of the generated images. These results are consistent with our goals of using them mentioned in the introduction.

\subsection{Discussions}

    \textbf{The Privacy Issues.}
    As mentioned in~\cite{shao2023survey}, the transmission of features is one of the existing ways of information sharing in FL. And considering the amount of data uploaded from clients, our method exhibits significantly lower privacy leakage compared to other FL methods.
    Moreover, to explore the feasibility of recovering private information from SD using the noise-added features, we focus on validating whether the generated images contain any privacy-sensitive information, such as text, faces, etc. For example, in some images with private text, we claim that the generated images with completely different texts do not leak the user's privacy, even though they appear to be similar. In Figure~\ref{SupVis_privacy}, we select four categories from OpenImage that may involve privacy-sensitive information and show some client images and their corresponding generated images. These results demonstrate that FedDISC has a low risk of leaking privacy-sensitive information during generation. Due to the space limitation, we provide further discussions in supplementary materials. 

   \textbf{Communication and Computational Complexity.}
      In Table~\ref{commu}, to compare communication and computational complexity with other compared methods, we conduct statistical analyses on DomainNet. Among the compared methods requiring iterations, the uploading and downloading of the image encoder is needed in each round. Ceiling needs to upload all client images. In contrast, FedDISC solely requires one time of downloading image encoder and uploading image features, incurring negligible communication. 
    
    Concerning computational complexity, since the server is generally not constrained by device performance due to tasks involving client scheduling and model aggregation, we only assess computation on the clients. All computation of Ceiling proceeds on the server and is devoid of reference value. Among the compared methods requiring iterations, multiple rounds of backpropagation exhibit more computing. But FedDISC needs a single forward propagation for feature extraction, resulting in approximately one-third of the computation compared to backpropagation. The aforementioned experiments demonstrate that FedDISC holds significant advantages in terms of communication and client computational complexity, underscoring its practicality.

    \section{Conclusion}
    
    In this paper, we explore the task of one-shot semi-FL and propose FedDISC, a new method that integrates pre-trained DMs into the semi-FL framework for the first time. In a single communication round and without any client training, our method achieves performance comparable to the ceiling performance and even surpasses it in some cases. The introduction of domain representations and clustering centroids further enhances the quality and stability of generation. Extensive quantitative and visualization experiments demonstrate the excellent performance of our method and underscore the potential and prospects of DMs within the FL.
\section{Acknowledgements}
This work was supported in part by the National Key R\&D Program of China (No.2021ZD0112803), the National Natural Science Foundation of China (No.62176061), STCSM project (No.22511105000), the Shanghai Research and Innovation Functional Program (No.17DZ2260900), and the Program for Professor of Special Appointment (Eastern Scholar) at Shanghai Institutions of Higher Learning.

\bibliography{references}

\end{document}